\documentclass{article}

\usepackage{arxiv}

\usepackage[utf8]{inputenc} 
\usepackage[T1]{fontenc}    
\usepackage{hyperref}       
\usepackage{url}            
\usepackage{booktabs}       
\usepackage{amsfonts}       
\usepackage{nicefrac}       
\usepackage{microtype}      
\usepackage{lipsum}
\usepackage{graphicx}
\graphicspath{ {./images/} }
\usepackage{pifont}
\usepackage{amssymb}
\usepackage{xcolor}
\usepackage{tabularx}
\usepackage{multirow}  
\usepackage{booktabs}
\usepackage{xcolor} 
\usepackage{url}
\usepackage[misc]{ifsym}
\usepackage{subcaption}
\newcommand\blfootnote[1]{%
  \begingroup
  \renewcommand\thefootnote{}\footnote{#1}%
  \addtocounter{footnote}{-1}%
  \endgroup
}

\title{Revisiting Network Perturbation for Semi-Supervised Semantic Segmentation}

\author{
 Sien Li \\
  College of Computer and Data Science\\ Fuzhou University\\
  Fuzhou 350108, China\\
  \texttt{LlistenL@163.com} \\
   \And
 Tao Wang \\
  School of Computer and Big Data\\
  Minjiang University\\
  Fuzhou 350108, China \\
  \texttt{twang@mju.edu.cn} \\
  \And
 Ruizhe Hu \\
    School of Computer Science and Mathematics\\
    Fujian University of Technology\\
    Fuzhou 350118, China \\
  \texttt{lkoor225@gmail.com} \\
  \And 
   Wenxi Liu \\
  College of Computer and Data Science\\ Fuzhou University\\
  Fuzhou 350108, China\\
  \texttt{wenxi.liu@hotmail.com} \\
}

\begin{document}
\maketitle
\begin{abstract}
In semi-supervised semantic segmentation (SSS), weak-to-strong consistency regularization techniques are widely utilized in recent works, typically combined with input-level and feature-level perturbations. However, the integration between weak-to-strong consistency regularization and network perturbation has been relatively rare. We note several problems with existing network perturbations in SSS that may contribute to this phenomenon. By revisiting network perturbations, we introduce a new approach for network perturbation to expand the existing weak-to-strong consistency regularization for unlabeled data. Additionally, we present a volatile learning process for labeled data, which is uncommon in existing research. Building upon previous work that includes input-level and feature-level perturbations, we present MLPMatch (Multi-Level-Perturbation Match), an easy-to-implement and efficient framework for semi-supervised semantic segmentation. MLPMatch has been validated on the Pascal VOC and Cityscapes datasets, achieving state-of-the-art performance. Code is available from~\url{https://github.com/LlistenL/MLPMatch}.\blfootnote{Paper accepted by the 7th Chinese Conference on Pattern Recognition and Computer Vision (PRCV 2024). \underline{\textit{Citation:}} Li, S., Wang, T., Hu, R., Liu, W. (2025). Revisiting Network Perturbation for Semi-supervised Semantic Segmentation. In: Lin, Z., et al. Pattern Recognition and Computer Vision. PRCV 2024. Lecture Notes in Computer Science, vol 15042. Springer, Singapore. https://doi.org/10.1007/978-981-97-8858-3\_11}
\end{abstract}


\section{Introduction}
Semantic segmentation has been a long-standing and fundamental problem in computer vision, and recent years witnessed significant performance improvements because of the impressive ability of deep neural networks to capture long-range and high-order interactions among local features~\cite{long2015fully,zhao2017pyramid,chen2018encoder,xie2021segformer}. One of the key challenges in semantic segmentation, however, is that the task requires a laborious and time-consuming manual labeling process on large datasets for models to learn well. Unlike image annotation, image acquisition is comparatively easier, prompting the rise of Semi-supervised Semantic Segmentation (SSS). This method aims to train segmentation models using limited labeled data and a large pool of unlabeled data, garnering considerable interest in recent times.
In terms of the paradigms proposed for SSS, mainstream methods have evolved from GAN-based adversarial training~\cite{souly2017semi} to self-training-based~\cite{ST++,cics-r} and the widely-adopted consistency regularization framework~\cite{ouali2020semi,french2020semi,chen2021semi,zou2020pseudoseg}. 
In particular, weak-to-strong consistency regularization expects a model to make consistent predictions on strongly perturbed and weakly perturbed versions of the same input image.
Intuitively, the model is more likely to generate high-quality predictions on the weakly perturbed images, while strong perturbations introduce additional variations, and the learning to generate consistent predictions on these inputs helps avoid systematic cognitive bias. 

\begin{figure}[t]
    \centering
    \includegraphics[trim={3.6cm 1.0cm 3cm 0.6cm},clip,width=0.5\textwidth]{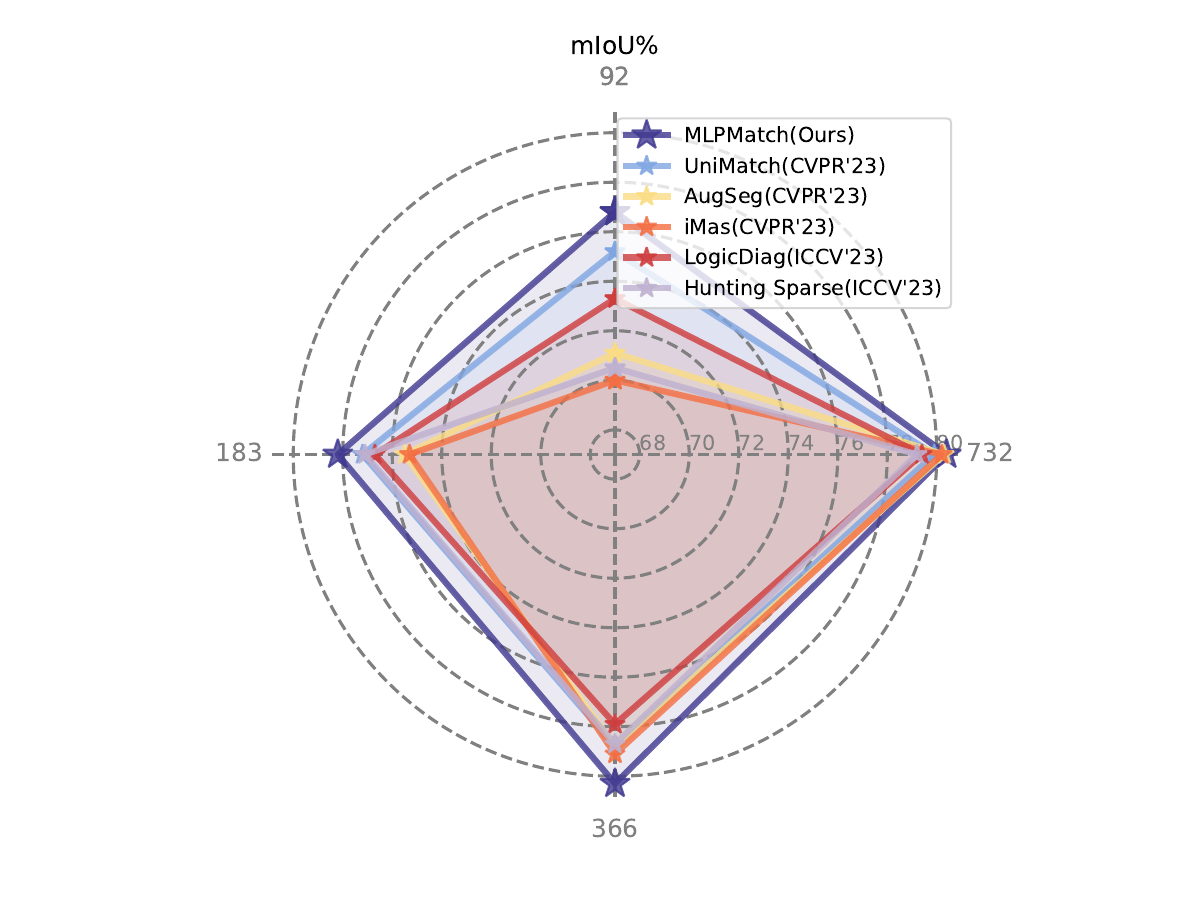}
    \caption{Performance comparison of MLPMatch with recent SOTA methods on the Pascal VOC dataset using a DeepLabv3plus and ResNet-101 model. The radii correspond to different quantities of labeled images.}
    \label{fig:compare_sota}
    \vspace{-4mm}
\end{figure}

It should be noted that it is non-trivial to design effective perturbation strategies for consistency regularization. Most recent work considers perturbations either at the input level or the feature level, or both.
AugSeg~\cite{augseg}, for instance, broadens the conventional augmentation repertoire with an intensity-based random augmentation strategy. Moreover, AugSeg adjusts the likelihood of applying CutMix based on the complexity of individual instances, rather than treating each instance uniformly.
iMas~\cite{imas} adopts an approach of assessing the difficulty of different instances at different time steps to apply varying degrees of augmentation. 
This helps to avoid the model encountering over-challenging augmentations early on and prevents excessive augmentation on difficult images, which could hinder effective learning by the model. UniMatch~\cite{unimatch}, on the other hand, performs perturbations at the feature level and the dual input level.

Another possibility is to perform network perturbation, i.e., perturbation induced by network parameters or architecture. For example, GCT~\cite{ke2020guided} and CPS~\cite{cps} train two identical networks with different initialization, while enforcing consistency by using one network to supervise the other. Likewise, Dual Student~\cite{ke2019dual} uses a bidirectional stabilization constraint between two models. We note that there are several problems in the existing network perturbation methods: (1) Compared to input or feature perturbation, existing network perturbation methods require at least two networks, which is computationally expensive and more suited towards the co-training framework. (2) These methods typically use supervision provided by networks with different initializations or architectures to increase the likelihood of overcoming cognitive bias caused by limited labeled data, that is insufficient to represent the intra-class diversity. In addition, perhaps more fundamental is that these methods cannot extract features that are sufficiently diverse from the same image to execute weak-to-strong consistency regularization, which is possible with both input and feature level perturbation. (3) Another common trait in the existing network perturbation methods is that it is only applied on unlabeled data, hence it is unclear whether we could reap similar benefit from labeled data.

In order to solve the problems above, we revisit network perturbation for semi-supervised semantic segmentation in this work. Inspired by the seminal work on dropping network layers~\cite{huang2016deep,larsson2016fractalnet}, we propose a simple yet effective strategy for network perturbation, MLPMatch, that is both computationally efficient and surprisingly effective.
The key idea of MLPMatch is deactivating layers of a DNN randomly. In other words, some layers in the network are skipped and not activated during forward propagation, resulting in a weakened version of the network. 
For example, switching off a certain convolutional block while retaining the skip connection can be easily implemented in most networks because of the extensive use of skip connections. The benefit of our proposed method is threefold: (1) It can be easily implemented with a single network without additional parameter storage requirements. (2) It can extract differentiated features from same image to perform weak-to-strong consistency regularization, and it does not fail to bring benefits due to the combination with rich perturbations at other levels of consistency, as demonstrated in our subsequent experiments.
(3) We show later that, network perturbation can also be applied on labeled data through a volatile learning process to obtain additional performance gains.

More specifically, we apply network perturbation to both unlabeled and labeled data in our approach. For unlabeled data, we perform forward propagation on weakly augmented inputs to obtain prediction for weakened versions of the network with certain convolutional blocks deactivated. We can therefore perform weak-to-strong consistency regularization between it and the prediction from a normal network. For labeled data, we introduce a volatile learning of labeled image to mitigate the adverse effect resulting from the difference in fitting progress between labeled data and unlabeled data. We conduct extensive experiments to verify the efficacy of the proposed method. In particular, based on UniMatch which covers weak-to-strong consistency at input and feature levels, our Multi-Level-Perturbation approach, MLPMatch, incorporates network perturbation and achieves state-of-the-art performance on Pascal VOC and Cityscapes datasets. Our main contributions are as follows:

\begin{itemize}
    \item We propose a simple, effective, and computationally efficient approach to integrate network perturbation into weak-to-strong consistency.
    \item We introduce a volatile learning process to counteract the negative impact caused by discrepancies in the uneven rates of fitting progress between labeled data and unlabeled data.
    \item Our approach, MLPMatch, achieves state-of-the-art performance on the Pascal VOC and the Cityscapes benchmark datasets.
\end{itemize}

\section{Related Work}

\subsection{Semi-supervised Semantic Segmentation}
In recent years, semi-supervised semantic segmentation has made remarkable progress, with a variety of new methods emerging, such as: semi-supervised contrastive learning methods~\cite{protoconsis}, segmentation prediction based on prototypes~\cite{protoview}, training methods based on heterogeneous frameworks like CNN and ViT~\cite{semicvt}, and so on. In particular, most mainstream approaches currently utilize a combination of pseudo-labeling techniques and consistency regularization in training, and their presence can be found in the methods mentioned above.
In terms of pseudo-labeling techniques, the family of methods based on confidence threshold filtering proposed in FixMatch~\cite{FixMatch} and its variants is widely embraced for its simplicity and effectiveness. In terms of consistency regularization, weak perturbations at the input level such as flipping and resizing, as well as strong perturbations like blurring and grayscale adjustments, are commonly used in SSS training~\cite{ST++,augseg,imas,unimatch,hunting,cics-r,ccvc}. Moreover, perturbations at the feature level, such as random channel dropout, are also extensively utilized.

Many recent works have delved deeper into this domain. For instance, iMas~\cite{imas} considers the instance's difficulty for varying degrees of perturbation at the input level, while AugSeg~\cite{augseg} expands the repertoire of input-level perturbations and proposes intensity-based variations. Furthermore, UniMatch~\cite{unimatch} integrates both input-level and feature-level perturbations. In contrast, this work primarily explores network perturbations that are less conventional in SSS. 

\subsection{Network Perturbation}
In terms of existing SSS methods with network perturbation, CPS~\cite{cps} exemplifies network perturbation by training two networks with differing initializations to enforce consistency between their predictions. CPS primarily benefits from using predictions of another dissimilar subnet as supervisory signals, enhancing its ability to mitigate inherent cognitive biases. However, unlike input and feature-level perturbations, these methods may struggle to extract diverse features from the same image due to the inherent homogeneity of the two networks, which is crucial for enforcing consistency constraints.
To address this issue, our work focuses on exploring network perturbations tailored for integration with consistency regularization, aiming to unlock their potential in SSS.

\section{Our Approach}
In this section, we describe the proposed MLPMatch framework in detail. Specifically, we first present a brief introduction to the semi-supervised semantic segmentation (SSS) task. This is followed by detailed descriptions of our approach for both unlabeled and labeled images.

\subsection{Preliminaries}
\label{sec:method:overview}
In semi-supervised semantic segmentation, we are given a small set of labeled images $\mathcal{D}_x = \{(\mathbf{x}_i,\mathbf{y}_i)\}_{i=1}^{M_x}$, where $\mathbf{x}_i \in \mathcal{X}$ and $\mathbf{y}_i \in \mathcal{Y}$ are the $i$-th image and the corresponding pixelwise ground-truth label, and $M_x$ is the number of images in the set of labeled images. In addition, we have a large set of unlabeled images $\mathcal{D}_u = \{(\mathbf{u}_i)\}_{i=1}^{M_u}$, where $\mathbf{u}_i \in \mathcal{X}$ is the $i$-th unlabeled image, and $M_u$ is the number of images in the set of unlabeled images. Our goal is to train a segmentation model $\mathcal{R}$ that maps any input image to the label space, i.e., $\mathcal{R}: \mathcal{X} \rightarrow \mathcal{Y} $.
In our work, $\mathcal{X} = \mathbb{R}^{H\times W\times 3}$ indicates that both labeled images and unlabeled images have a spatial dimension of $H \times W$ and $3$ channels. And $\mathcal{Y} = \mathcal{C}^{H\times W}$ are the spatial dimension of both ground-truth labels and pseudo-labels, where $\mathcal{C}=\{ 1 \dots C \}$ is the set of $C$ classes. 

Typically, $\mathcal{R}$ includes an encoder $\mathcal{F}(\cdot)$ to obtain a latent feature representation $\mathbf{z}_i \in \mathbb{R}^{\hat{H} \times \hat{W} \times D}$ from $\mathbf{x}_i$ or $\mathbf{u}_i$, and a decoder $\mathcal{G}(\cdot)$ to further convert $\mathbf{z}_i$ into $\mathbf{p}_i \in \mathbb{R}^{H \times W \times C}$.

Given a labeled image $\mathbf{x}_i$ and its corresponding ground-truth $\mathbf{y}_i$, the cross-entropy loss $\mathcal{L}_x$ is calculated between the prediction under the weakly perturbed input view $A^w(\mathbf{x}_i)$ and $\mathbf{y}_i$. Loss $\mathcal{L}_x$ can be written as:

\begin{equation}
\small
\mathcal{L}_{x} = \frac{1}{M_x} \sum_{i=1}^{M_x}\frac{1}{H\times W}\sum_{j=1}^{H\times W} \mathcal{L}_{ce}\Big(p_{ij}(A^w(\mathbf{x}_i)),y_{ij}\Big)
\label{eq:suploss}
\end{equation}

\noindent where $p_{ij}(\cdot)$ and $y_{ij}$ refer to the prediction and the ground-truth label of the $i$-th image at the $j$-th site, respectively. For unlabeled images $\mathbf{u}_i$, the predictions from $A^w(\mathbf{u}_i)$ will be used as pseudo-labels $\mathbf{y}^{u}_i$ through an argmax operation and a high-confidence threshold filtering for predictions with strong perturbations.
The consistency regularization at the input level is applied between predictions of unlabeled images with weak input-level perturbations and unlabeled images with strong input-level perturbations. Additionally, our baseline method applies two random strong input-level perturbations, $A^{s1}$ and $A^{s2}$, to the unlabeled data and calculates $\mathcal{L}_{u}^{s1}$ and $\mathcal{L}_{u}^{s2}$. For brevity, we combine them into $\mathcal{L}_{u}^{s}$ as:

\begin{equation}
\small
\mathcal{L}_{u}^s  = \frac{1}{2M_u} \sum_{i=1}^{M_u}\frac{1}{H\times W}\sum_{j=1}^{H\times W}  \mathcal{L}_{ce}\Big(p_{ij}(A^{s1/2}(\mathbf{u}_i)), y_{ij}^u\Big) \cdot \mathcal{M}_i
\label{eq:unsuploss}
\end{equation}

\noindent where $\mathcal{M}_i$ is a binary mask that is set to 1 when $p_{ij}(A^w(\mathbf{u}_i)) > \tau$, and 0 otherwise.

Besides, our baseline method performs consistency regularization with feature-level perturbation $A^{fp}$. Specifically, channel-wise dropout is applied to $\mathcal{F}(A^w(\mathbf{u}_i))$ with $\mathbf{y}_i^u$ as the supervision. So the corresponding loss term $\mathcal{L}_u^{fp}$ is written as:
\begin{equation}
\small
    \mathcal{L}_{u}^{fp} = \frac{1}{M_u} \sum_{i=1}^{M_u}\frac{1}{H\times W}\sum_{j=1}^{H\times W} \mathcal{L}_{ce}\Big(p_{ij}(A^{fp}(\mathbf{u}_i), y^u_{ij}\Big) \cdot \mathcal{M}_i 
\label{eq:fploss}
\end{equation}



\begin{figure*}[t]
    \centering
    \includegraphics[trim={0cm 5cm 5.5cm 0cm},clip,width=\textwidth]{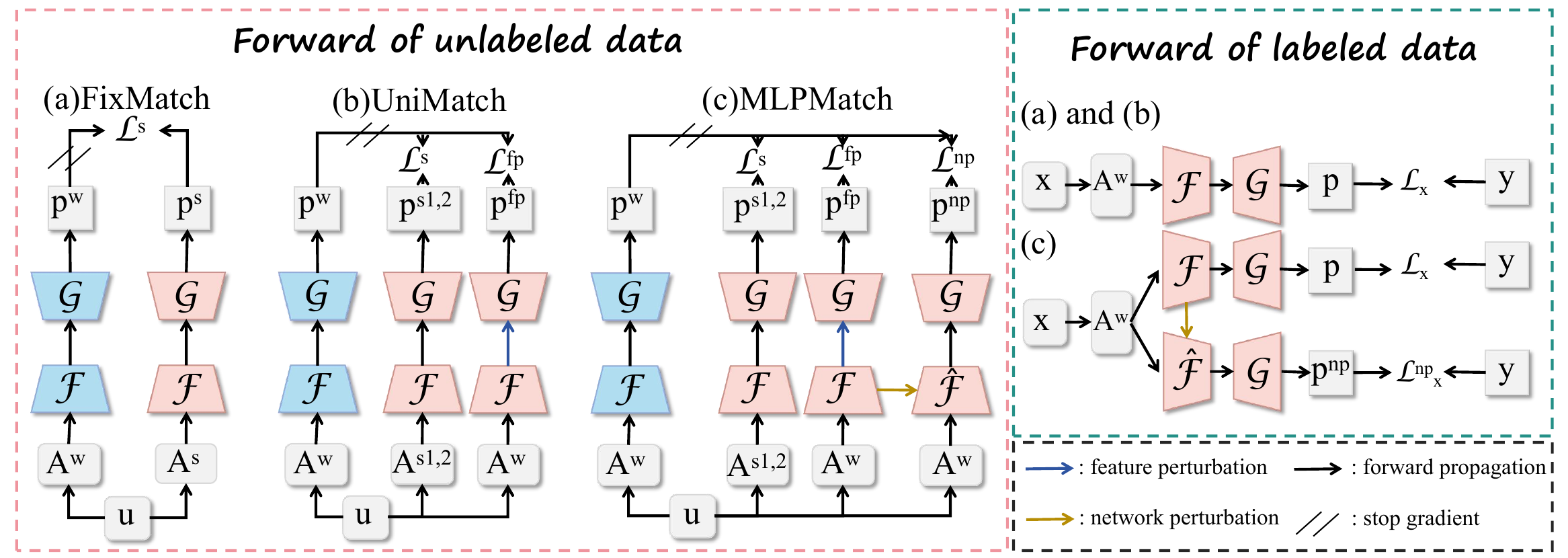}
    
    \caption{An overview to MLPMatch. See comparisons with (a) FixMatch and (b) UniMatch in the training of unlabeled data on the left side. See comparisons with (a) FixMatch and (b) UniMatch in the training of labeled data on the right side. $A^w$ and $A^s$ represent the weak and strong input-level perturbations (augmentations).}
    \label{fig:DropPath}
\end{figure*}


\begin{figure}[t]
    \centering
    \includegraphics[trim={0cm 13.5cm 11cm 0.5cm},clip,width=0.9\textwidth]{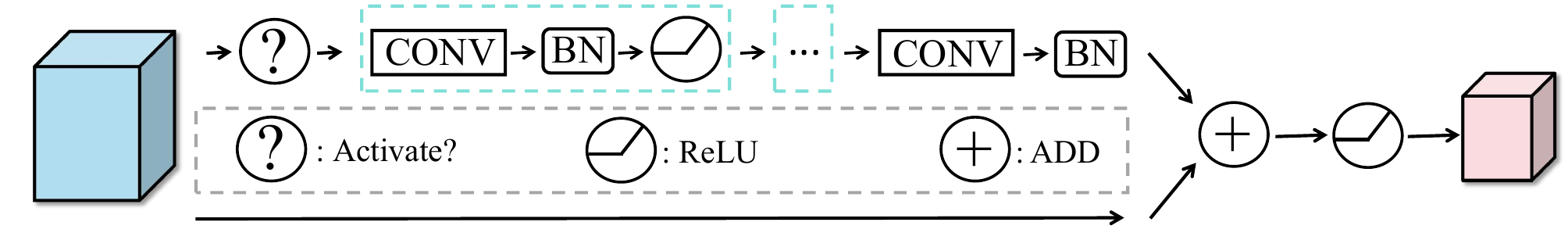}
    
    \caption{Details of the BottleNeck in $\mathcal{\hat{F}}$. The convolutional forward pass of the BottleNeck is randomly activated or deactivated.}
    \label{fig:bottleneck}
    \vspace{-4mm}
\end{figure}


\subsection{Weak-to-strong consistency based on network perturbation}
As already discussed, recent work~\cite{augseg,unimatch,imas,FixMatch} uses the weak-to-strong consistency to force the decision boundary of a model to be located in areas with low sample density. However, most works are limited to the input-level and feature-level perturbations. In this paper, we introduce Network Perturbation (NP) to form a weak version of the encoder $\mathcal{\hat{F}}$ from the original $\mathcal{F}$. More specifically, the convolutional forward of bottlenecks will be deactivated randomly to form $\mathcal{\hat{F}}$ from $\mathcal{F}$, as shown in Fig.~\ref{fig:bottleneck}.
  
Based on the idea that over-strong perturbations negatively impact model performance~\cite{unimatch}, we limit only one bottleneck layer in the entire $\mathcal{\hat{F}}$ to execute NP.
Though $\mathcal{\hat{F}}$ and $\mathcal{F}$, we can obtain diverse feature representations from the same image. We then perform weak-to-strong consistency between $\mathcal{G}(\mathcal{\hat{F}}(\mathbf{u}_i))$ and $\mathcal{G}(\mathcal{F}(\mathbf{u}_i))$.
Specifically, the weak-to-strong consistency loss can be written as:
\begin{equation}
\small
\mathcal{L}_{u}^{np} = \frac{1}{M_u} \sum_{i=1}^{M_u}\frac{1}{H\times W}\sum_{j=1}^{H\times W}\mathcal{L}_{ce}\Big(\hat{p}_{ij}(A^w(\mathbf{u}_i)), y^u_{ij}\Big) \cdot \mathcal{M}_i
\label{eq:nploss}
\end{equation}
Here, $\hat{p}_{ij}$ represents the prediction of $\mathbf{u}_{ij}$ using the encoder $\mathcal{\hat{F}}$.

\subsection{A volatile learning process for labeled data}
Recent work~\cite{FixMatch} has introduced a high-confidence threshold (\textit{e.g.,}~$0.95$) to filter noise in pseudo-labels. This implies that in the early stages of training, when few pseudo-labels pass the filtering process, the model relies primarily on labeled data.
In the scenario of semi-supervised training, the limited labeled data cannot cover the diversity in data distribution within different classes. As a result, this means that we risk fitting the labeled data to a much greater extent than fitting the unlabeled data.
Additionally, the over-confidence phenomenon further results in incorrect pseudo-labels passing through the filtering process and contributing to training. To address this issue, we propose a strategy of perturbing supervised learning data, which is uncommon in previous work. We refer to this practice as volatile learning and expect that this ``volatility'' can alleviate the impact of fitting labeled data far ahead of fitting unlabeled data.
Specifically, we utilize both $\mathcal{\hat{F}}$ and $\mathcal{F}$ to fit labeled data, instead of using $\mathcal{F}$ alone. And the loss of volatile learning $\mathcal{L}_x^{np}$ can be written as:
\begin{equation}
\small
    \mathcal{L}_{x}^{np} = \frac{1}{M_x} \sum_{i=1}^{M_x}\frac{1}{H\times W}\sum_{j=1}^{H\times W} \mathcal{L}_{ce}\Big(\hat{p}_{ij}(A^w(\mathbf{x}_i)),y_{ij}\Big)
\label{eq:newsuploss}
\end{equation}

In addition, the difference in fitting progress between labeled and unlabeled data mentioned above intensifies as training progresses, so we set the relative weight of this loss term to a simple linear growth to reflect this trend.

\subsection{Our Holistic Framework: MLPMatch}
In summary, we utilize network perturbations on both labeled data and unlabeled data to enhance model performance. From Fig.~\ref{fig:DropPath}, we can clearly see the comparisons between our baseline methods and MLPMatch. The total loss of MLPMatch can be written as:
\begin{equation}
\small
    \mathcal{L}_{total} = (\lambda_x - \lambda_x^{np})\mathcal{L}_x + \lambda_x^{np}\mathcal{L}_x^{np} + \lambda_u^{s}\mathcal{L}_u^s + \lambda_u^{fp}\mathcal{L}_u^{fp} + \lambda_u^{np}\mathcal{L}_u^{np}
\label{eq:totalloss}
\end{equation}
Following our baseline, $\lambda_u^s$ is set to 0.5, and we set $\lambda_u^{fp}$ and $\lambda_u^{np}$ to 0.25 to maintain consistent loss weights between labeled and unlabeled data. $\lambda_x^{np}$ is a linear increasing value starting from 0. In addition, we conduct experiments with different $\lambda_x^{np}$ in the experiments section.

\section{Experiments}
In this section, we first introduce our experimental settings. Next, we compare our method with recent state-of-the-art (SOTA) methods in SSS. In particular, it should be noted that our method did not introduce any modifications during model inference, as compared to the baseline methods.
\begin{table}[t]
	\centering
	\begin{tabular}{r|c|c c c c}
		\toprule
  \midrule
		\multirow{2.5}{*}{\textbf{Pascal VOC}} & \multirow{2.5}{*}{\textbf{Resolution}}
        & \multicolumn{4}{c}{ResNet-101} \\
		
		\cmidrule{3-6}
		~ & ~ & 92 & 183 & 366 & 732  \\
        
		\midrule
            SupOnly & 321×321 & 45.1 & 55.3 & 64.8  &  69.7 \\
        \midrule
            CPS\ \tiny{\textcolor{gray}{[CVPR'21]}}\cite{cps} & 512×512 & 64.1 & 67.4 & 71.4 & 75.9 \\
            ST++\ \tiny{\textcolor{gray}{[CVPR'22]}} \cite{ST++}& 321×321 & 65.2 & 71.0 & 74.6 & 77.3 \\
            U$^2$PL\ \tiny{\textcolor{gray}{[CVPR'22]}} \cite{U2PL}&512×512 & 68.0 & 69.2 & 73.7 & 76.2 \\
            PS-MT\ \tiny{\textcolor{gray}{[CVPR'22]}}\cite{psmt}&512×512 & 65.8 & 69.6 & 76.6 & 78.4 \\
            PCR\ \tiny{\textcolor{gray}{[NeurIPS'22]}}\cite{pcr}&512×512 & 70.1 & 74.7 & 77.2 & 78.5 \\
            ESL\ \tiny{\textcolor{gray}{[ICCV'23]}}\cite{esl}&513×513 & 70.1 & 74.1 & 78.1 & 79.5 \\
            SemiCVT\ \tiny{\textcolor{gray}{[ICCV'23]}}\cite{semicvt}&513×513 & 68.6 & 71.2 & 75.0 & 78.5 \\
            Hunting Sparse\ \tiny{\textcolor{gray}{[ICCV'23]}}\cite{hunting}&513×513 & 70.5 & 77.1 & 78.7 & 79.2\\
            LogicDiag\ \tiny{\textcolor{gray}{[ICCV'23]}}\cite{logic}&513×513 & 73.3  & 76.7& 77.9 & 79.4 \\
            3-cps\ \tiny{\textcolor{gray}{[ICCV'23]}}\cite{dcps}&321×321 & 75.4 & 76.8 & 79.6 & 80.4\\
            3-cps\ \tiny{\textcolor{gray}{[ICCV'23]}}\cite{dcps}&513×513 & 75.7 & 77.7 & 80.1 & \textbf{80.9} \\
            iMas\ \tiny{\textcolor{gray}{[CVPR'23]}}\cite{imas}&513×513 & 70.0 & 75.3 & 79.1 & 80.2 \\
            augseg\ \tiny{\textcolor{gray}{[CVPR'23]}}\cite{augseg}&513×513 & 71.1 & 75.5 & 78.8 & 80.3 \\
            UniMatch\ \tiny{\textcolor{gray}{[CVPR'23]}}\cite{unimatch}&321×321 & 75.2 & 77.2 & 78.8 & 79.9  \\
            \midrule
            MLPMatch&321×321 & \textbf{76.8} & \textbf{78.2} & \textbf{80.3} & 80.4 \\
            \textcolor{blue}{Gain } & - & \textcolor{blue}{↑31.7} &
            \textcolor{blue}{↑22.9} &
            \textcolor{blue}{↑15.5}&
            \textcolor{blue}{↑10.7}\\

    \midrule
  \bottomrule
	\end{tabular}
     \vspace{2mm}
	\caption{Comparison with SOTAs on the \textbf{Pascal VOC} dataset. The experiments were conducted with a DeepLabV3+ and ResNet-101 model. The numbers in the second row indicate the amount of labeled data available for a particular partition protocol.}
	\label{table:pascal_r101}
\end{table}

\begin{table}[h]
	\centering
	\begin{tabular}{r|c|c c c c}
		\toprule
  \midrule
		\multirow{2.5}{*}{\textbf{Pascal VOC}} & \multirow{2.5}{*}{\textbf{Resolution}}
        & \multicolumn{4}{c}{ResNet-50} \\
		
		\cmidrule{3-6}
		~ & ~ & 92 & 183 & 366 & 732  \\
		\midrule
            SupOnly & 321×321 & 44.0 & 52.3 & 61.7  &  66.7  \\
        \midrule
            3-cps\ \tiny{\textcolor{gray}{[ICCV'23]}}\cite{dcps}&513×513  & 73.1 & 74.7 & 77.1 & 78.8\\
            augseg\ \tiny{\textcolor{gray}{[CVPR'23]}}\cite{augseg}&513×513 & 64.2 & 72.2 & 76.2 & 77.4  \\
            UniMatch\ \tiny{\textcolor{gray}{[CVPR'23]}}\cite{unimatch}&321×321 & 71.9 & 72.5 & 76.0 & 77.4  \\
            \midrule
            MLPMatch&321×321 & \textbf{73.8} & \textbf{74.7} & \textbf{77.3} & \textbf{78.8} \\
            \textcolor{blue}{Gain } & - & \textcolor{blue}{↑29.8} &
            \textcolor{blue}{↑22.4} &
            \textcolor{blue}{↑16.4}&
            \textcolor{blue}{↑12.1}\\
        \midrule
  \bottomrule
	\end{tabular}
    \vspace{2mm}
	\caption{Comparison with SOTAs on the \textbf{Pascal VOC} dataset. The experiments were conducted with a DeepLabV3+ and ResNet-50 model.}
	\label{table:pascal_r50}
\end{table}

\vspace{-4mm}
\subsection{Experimental Setup}
\noindent\textbf{Datasets}. We conduct experiments on two widely used datasets, Pascal VOC 2012~\cite{pascal} and Cityscapes~\cite{cityscapes}, to demonstrate the efficacy of MLPMatch. Pascal VOC provides a semantic segmentation task with 21 classes including the background class. This dataset contains 1464 high-quality annotated images and 1449 validation images. Following~\cite{cps,ST++,U2PL}, 9118 additional coarsely-labeled images from the SBD dataset~\cite{hariharan2011semantic} have been added to the dataset. Ultimately, this resulted in a training set comprising 10582 images. Cityscapes~\cite{cityscapes} is an urban-oriented dataset consisting of 19 classes, with 2975 finely annotated training images and 500 validation images.

\noindent
\textbf{Training.} To compare fairly with previous works, extensive comparative experiments and ablation studies were conducted on the two datasets based on the DeeplabV3+~\cite{chen2018encoder} model with ResNet-50~\cite{he2016deep} or ResNet-101 backbone. 

 For \textbf{Pascal VOC 2012}, we trained for 80 epochs with an initial learning rate of 0.0010 and a Poly decay strategy, using an SGD optimizer and cross-entropy loss during training. The training resolution was set to 321. The mini-batch size is set to 16. The $\tau$ was set to 0.95.

For \textbf{Cityscapes}, we trained for 240 epochs with an initial learning rate of 0.0050 and a Poly decay strategy, using SGD optimizer and OHEM loss during training. The training resolution was set to 769. The mini-batch is set to 16.  The $\tau$ was set to 0.00.

\noindent\textbf{Evaluation.} We followed the same partitions used in UniMatch~\cite{unimatch} for a fair comparison. Following recent works~\cite{unimatch,imas,augseg,dcps}, Sliding window inference is adopted with the Cityscapes dataset. The evaluation metric for all experiments is mIoU (mean Intersection over Union).

\begin{table*}[t]

    \centering
    \begin{tabular}{ccc|cccc}
        \toprule
        \midrule
          Baseline & $\mathcal{L}_{u}^{np}$ & $\mathcal{L}_{x}^{np}$ & 92 & 183 & 366 & 732 \\
          \midrule
          \ding{51} & &  & 71.9 / 75.2 & 72.5 / 77.2 & 76.0 / 78.8 & 77.4 / 79.9  \\
          \ding{51} &\ding{51} & &73.2 / 75.2 & \textbf{75.2 / 78.3} & 76.6 / 79.1 & 78.8 / 80.1\\
          \ding{51} &\ding{51} & \ding{51} & \textbf{73.8 / 76.8 } & 74.7  / 78.2  & \textbf{77.3 / 80.3 } & \textbf{78.8  / 80.4 }\\
          \midrule
        \bottomrule
    \end{tabular}
    \vspace{2mm}
    \caption{Ablations on different components of MLPMatch. Results are obtained on \textbf{Pascal VOC} with a DeepLabv3+ and ResNet-50 / ResNet-101 model.}
    \label{tab:abl:component}
\end{table*}

\begin{table}[h]
\centering
        \begin{tabular}{r|ccc}
    		\toprule
      \midrule
    		\multirow{2.5}{*}{\textbf{Cityscapes}} & \multicolumn{3}{c}{ResNet-50}  \\
    		
    		\cmidrule{2-4}
    		~ &1/32 (183)& 1/16 (372) & 1/4 (744) \\
    		
    		\midrule
    		
    		SupOnly & 57.4 &  63.3 & 73.1 \\
    	
    		ST++\ \tiny{\textcolor{gray}{[CVPR'22]}}\cite{ST++} & 61.4  & 73.8 & -  \\
                 
            U$^2$PL\ \tiny{\textcolor{gray}{[CVPR'22]}}\cite{U2PL}& - & 70.6  & 77.2  \\

            UniMatch$^{\dag}$ \tiny{\textcolor{gray}{[CVPR'23]}}\cite{unimatch}& 69.2 & 73.9 & 77.2\\
    		
    		iMas \tiny{\textcolor{gray}{[CVPR'23]}}\cite{imas}& - & 74.3 & 78.1\\

            AugSeg \tiny{\textcolor{gray}{[CVPR'23]}}\cite{augseg}& - & 73.7 & \textbf{78.8}\\
        
      \midrule
    		MLPMatch\ \tiny{\textcolor{gray}{[OURS]}}  & \textbf{71.2} &\textbf{74.7} & 77.5\\
      
    		\textcolor{blue}{Gain } & \textcolor{blue}{↑13.8} &
            \textcolor{blue}{↑11.4} &
            \textcolor{blue}{↑4.4}\\
                \midrule
    		\bottomrule
    	\end{tabular}
    \vspace{2mm}
	\caption{Comparison with SOTAs on the \textbf{Cityscapes} dataset. The experiments were conducted with a DeepLabV3+ and ResNet-50 model. $\dag$ means our reproduced results with 769×769 resolution.}
	\label{table:cityscapes_all}
\end{table}

\subsection{Comparison with SOTAs}

\noindent \textbf{Pascal VOC 2012}. In Tab.~\ref{table:pascal_r101} and Tab.~\ref{table:pascal_r50}, we compare MLPMatch with recent SOTAs. It is evident that our approach significantly outperforms other methods when the quantity of labeled data is small (\textit{e.g.,} 92, 183, 366 partition protocols).

Tab.~\ref{tab:abl:component} provides a detailed demonstration of the performance improvements brought by our method under different partition protocols.

\begin{figure}[h]
    \centering
    \begin{minipage}{0.49\linewidth}
        \centering
        \includegraphics[trim={0.3cm 0.6cm 2.3cm 1.6cm},clip,width=\linewidth]{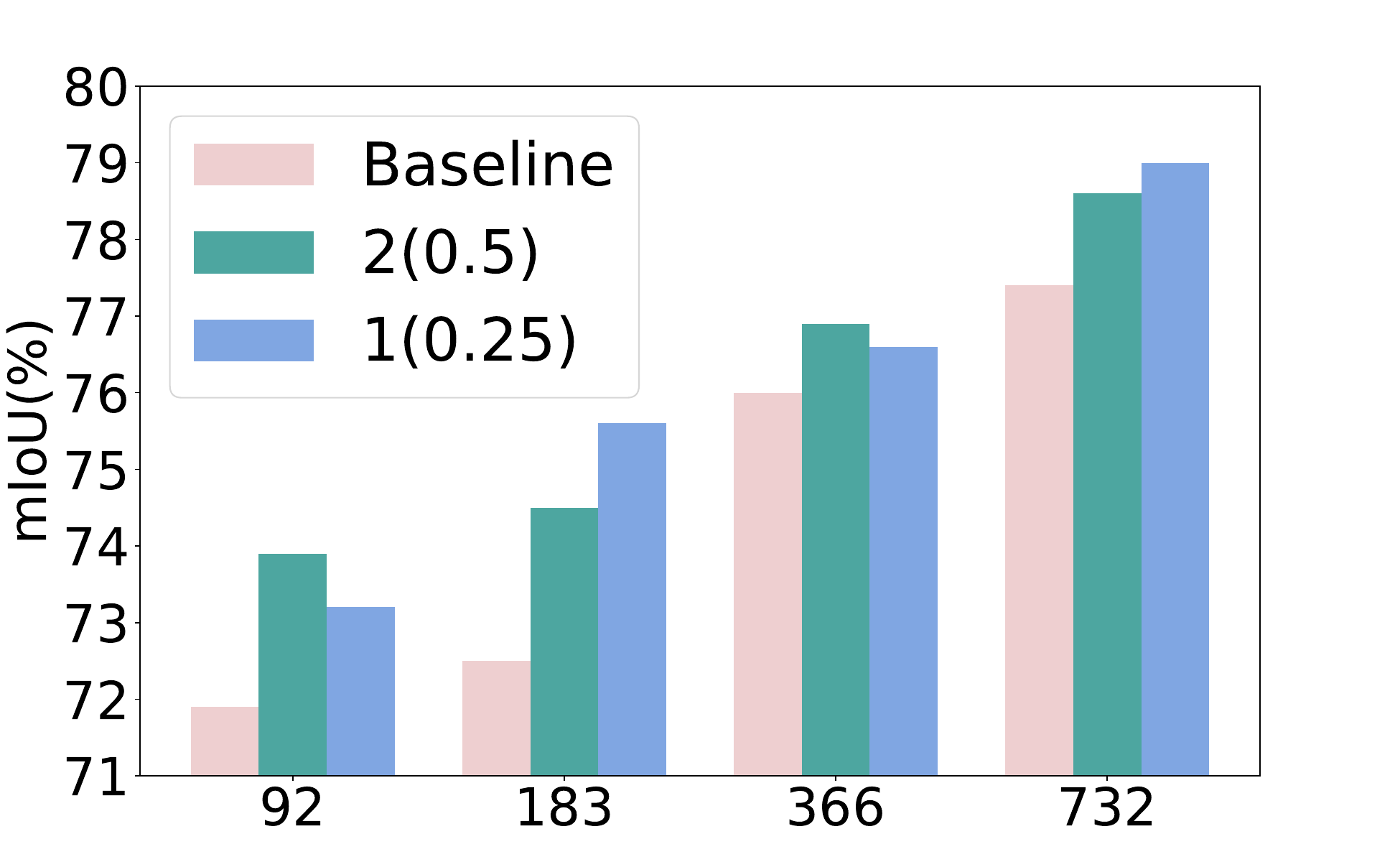}
        \subcaption{(a) Ablation - number of layers}
         \vspace{-2mm}
    \end{minipage}\hfill
    \begin{minipage}{0.49\linewidth}
        \centering
       \includegraphics[trim={0.3cm 0.6cm 2.3cm 1.6cm},clip,width=\linewidth]{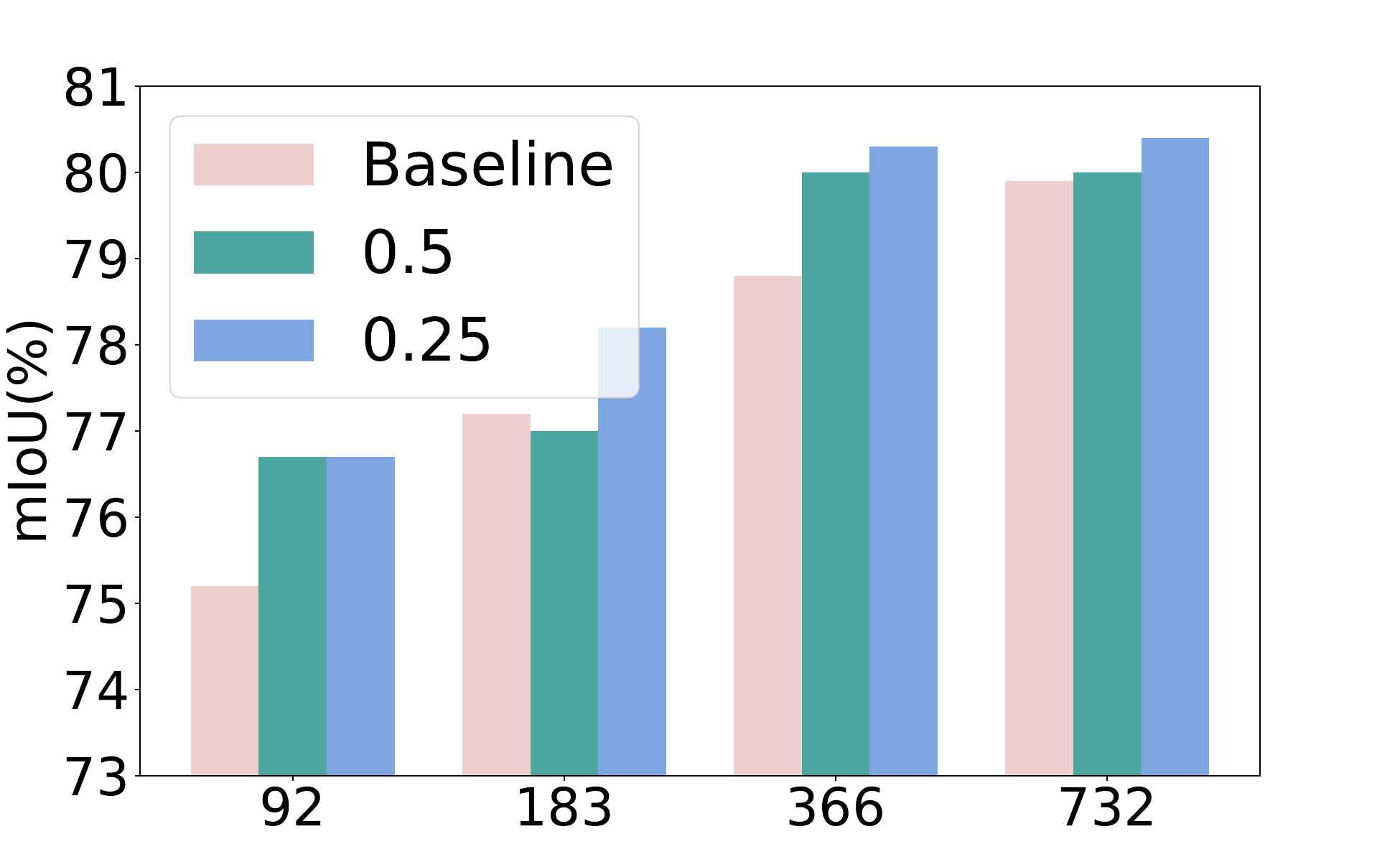}
         \subcaption{(b) Ablation - $\lambda_{x}^{np}$}
          \vspace{-2mm}
    \end{minipage}\hfill

\caption{More experimental results on (a) varying the number of layers for drop and (b) different values of $\lambda_x^{np}$. The experiments in (a) were conducted on the \textbf{Pascal VOC} dataset using a ResNet-50 backbone. The experiments in (b) were conducted on the \textbf{Pascal VOC} dataset using a ResNet-101 backbone.}
\label{fig:abl}
\end{figure}

\noindent \textbf{Cityscapes}. In Tab.~\ref{table:cityscapes_all}, we evaluate our method on the more challenging Cityscapes dataset, using ResNet-50 as the encoder. Compared to recent SOTA methods including UniMatch, iMas, and AugSeg, MLPMatch outperforms all methods under the 1/30 and 1/16 partition protocols. In particular, MLPMatch surpasses our baseline UniMatch by 2.0\% under the 1/32 partition protocol.



\begin{table}
    \centering
    \vspace{-4mm}
    \begin{tabular}{c|c c}
        \toprule
        \midrule
          $\lambda_{np}^x$ & 92 & 183  \\
          \midrule
          Linear increasing & \textbf{73.8} & \textbf{74.7} \\
          Fixed & 70.4 & 74.3\\
          \midrule
        \bottomrule
    \end{tabular}
    \vspace{2mm}
    \caption{Ablation studies on the choice of $\lambda_{np}^x$. Results were obtained on the \textbf{Pascal VOC} dataset using DeepLabv3+ and ResNet-50.}
    \label{tab:abl:npx}
\end{table}

\begin{figure}[h]
    \centering
    \includegraphics[trim={1.2cm 5.2cm 9.8cm 0.5cm},clip,width=0.8\textwidth]{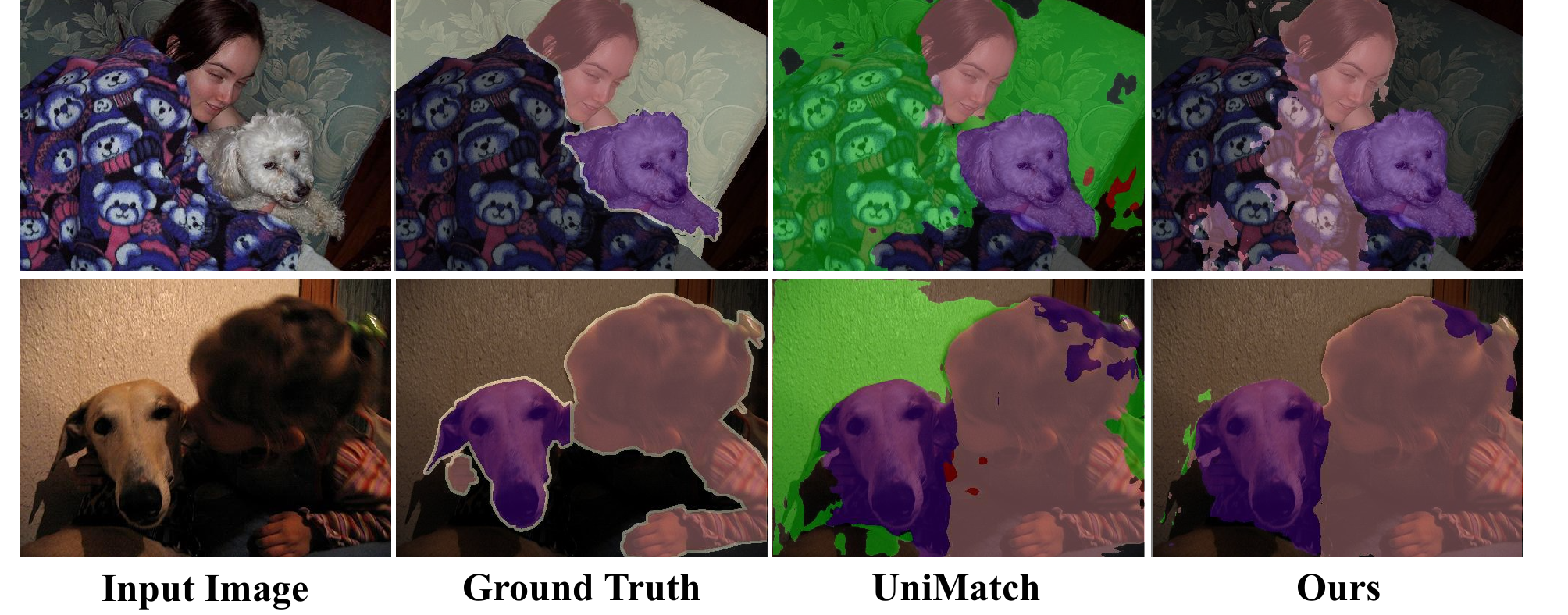}
    \caption{Qualitative comparisons on the \textbf{Pascal VOC} dataset. The DeepLabv3+ model with a ResNet-50 backbone was trained using the 92 labeled images partition protocol.}
    \label{fig:visualization}
    \vspace{-4mm}
\end{figure}

\begin{table}[t]
	\centering
	\begin{tabular}{r|c c |c c}
		\toprule
  \midrule
		\multirow{2.5}{*}{\textbf{Method}}
        & \multicolumn{2}{c}{WHU-CD~\cite{whu}} & \multicolumn{2}{c}{LEVIR-CD~\cite{levir}} \\
		
		\cmidrule{2-5}
		~ & 20\% & 40\%& 20\%& 40\%  \\
        
		\midrule
            SupOnly & 68.4 / 98.34 & 76.2 / 98.87 & 77.6 / 98.79  & 70.5 / 98.94\\
        \midrule
            UniMatch & 81.7 / 99.18 & 85.1 / 99.35 & 81.7 / 99.02 & 82.1 / 99.03 \\
            MLPMatch  & \textbf{85.5 / 99.38}& \textbf{86.2 / 99.40} & \textbf{82.8 / 99.07} & \textbf{83.0 / 99.07} \\
        \midrule
  \bottomrule
	\end{tabular}
     \vspace{2mm}
	\caption{Comparison with UniMatch on \textbf{Remote Sensing} tasks. Numbers in each cell denote the  changed-class IoU and overall accuracy, respectively. The fraction (\textit{e.g.,} 20\%) means the proportion of labeled images. All experiments were conducted using DeepLabv3+ and ResNet-50.}
	\label{table:remote}
\end{table}

\subsection{Ablations studies}
In this section, we perform a series of ablation studies on the Pascal VOC dataset.

\noindent \textbf{Effectiveness of each component}. From Tab.~\ref{tab:abl:component}, we can observe the efficacy of each component of MLPMatch. Specifically, the sequential addition of $\mathcal{L}_{u}^{np}$ and $\mathcal{L}_{x}^{np}$ gradually improves model performance, with our complete method providing the best mIoU.

\noindent  \textbf{Different number of bottleneck layer for NP.} In Fig.~\ref{fig:abl}(a), we experiment with implementing NP in varying numbers of bottleneck layers to achieve different levels of network perturbation. It is worth noting that the different numbers of NP operations in the last four layers of ResNet are achieved by setting various deactivation probabilities. The values in parentheses represent the deactivation probabilities for each layer in practice. It is evident that the model can achieve performance improvements through different levels of NP. In line with recent findings~\cite{unimatch}, which suggest that moderate perturbation is more beneficial for model learning, we implement NP once in $\mathcal{\hat{F}}$.

\noindent  \textbf{Different value of $\lambda_x^{np}$.} In Fig.~\ref{fig:abl}(b), we evaluate MLPMatch with different values of $\lambda_x^{np}$. It becomes clear that various values of $\lambda_x^{np}$ allow the model to benefit under almost all partition protocols. In our ablation study, a value of 0.25 outperforms 0.5; thus, for all our experiments, we set $\lambda_x^{np}$ to 0.25.

\noindent \textbf{Volatile learning of labeled data.} As shown in Tab.~\ref{tab:abl:npx}, using fixed weights in labeled perturbation learning does not lead to performance gains and may even be detrimental. However, employing  linearly increasing weights can provide significant benefits. We attribute this to the simple linear growth of the weight, which can roughly capture the increasing trend in fitting level differences between labeled and unlabeled data.

\noindent \textbf{Qualitative comparison of predictions.} From Fig.~\ref{fig:visualization}, we can see that even with very few labeled data (92), our approach, while not accurately predicting the fine details of the object boundaries, successfully maintains a rough outline of the objects without producing highly unreasonable predictions (the green category in the third column represents the ``train'' class).

\noindent \textbf{More domains.} We note that UniMatch has been validated for its effectiveness of their weak-to-strong consistency regularization in the remote sensing domain. Tab.~\ref{table:remote} presents a comparison between UniMatch and MLPMatch on remote sensing benchmarks.

\section{Conclusion}
In this paper, we present MLPMatch, an easy-to-implement and efficient framework for SSS. A key advantage of our method is its effectiveness and strong generalization. Compared to existing network perturbation methods in SSS, our approach can be seamlessly integrated with other weak-to-strong consistency regularization approaches to enhance model generalization. Additionally, we introduce perturbations in the learning process of labeled data, a strategy that has been relatively rare in previous research. MLPMatch achieves state-of-the-art performance on Pascal VOC and Cityscapes datasets. Furthermore, we validate its generalization and effectiveness in the remote sensing domain.
\newline\newline
\noindent \textbf{Acknowledgment.} This work is supported by the Fujian Natural Science Foundation~(2022J011112) and the Research Project of Fashu Foundation~(MFK23001).

\bibliographystyle{splncs04}
\bibliography{references}

\end{document}